\documentclass[cameraready]{Interspeech}

\title{CHUCKLE - When Humans Teach AI to Learn Emotions the Easy Way}

\author[affiliation={1}, orcid=0009-0002-6614-6742]{Ankush Pratap}{Singh}
\author[affiliation={1}, orcid=0000-0002-2310-7682]{Houwei}{Cao}
\author[affiliation={2}, orcid=0000-0001-9126-1430]{Yong}{Liu}

\address{
    $^1$ New York Institute of Technology, Department of Computer Science, New York, United States \\
    $^2$ New York University, Electrical and Computer Engineering Department, New York, United States
}

\email{asing213@nyit.edu, hcao02@nyit.edu, yongliu@nyu.edu}


\usepackage{comment}

\begin{document}

\maketitle

\begin{abstract}
Curriculum learning (CL) structures training from simple to complex samples, facilitating progressive learning. However, existing CL approaches for emotion recognition often rely on heuristic, data-driven, or model-based definitions of sample difficulty, neglecting the difficulty for human perception, a critical factor in subjective tasks like emotion recognition. We propose CHUCKLE (\textbf{C}rowdsourced \textbf{H}uman \textbf{U}nderstanding \textbf{C}urriculum for \textbf{K}nowledge \textbf{L}ed \textbf{E}motion Recognition), a perception-driven CL framework that leverages annotator agreement and alignment in crowd-sourced datasets to define sample difficulty, under the assumption that clips challenging for humans are similarly hard for neural networks. Experimental results suggest that CHUCKLE enhances the performance of  LSTMs and Transformers over non-curriculum baselines, while reducing the number of gradient updates, thereby enhancing both training efficiency and model robustness in both subject-dependent and subject-independent settings.

\end{abstract}

\section{Introduction}

Emotions shape the human experience, influencing communication, decision-making, and social interaction. Automatic emotion recognition seeks to infer human affective states from multi-modal signals such as speech \cite{GEORGE2024127015,schuller2018speech, lian2023survey}, text \cite{lian2023survey}, facial expressions \cite{lian2023survey, CANAL2022593}, gestures \cite{noroozi,gandi}, and physiological signals \cite{wenqian, wang}. Among these, speech emotion recognition (SER) remains central yet challenging, as it is hindered by speaker variability, contextual and cultural differences, noise, linguistic confounds, and the subjectivity of emotional perception. These factors make SER a high-variance, label-noisy problem where conventional training often fails to generalize. Curriculum learning (CL) offers a promising solution by structuring training from easy to hard samples, inspired by human learning \cite{cl_bengio}. In SER, this is well-motivated: starting with easy, unambiguous samples helps models learn stable low-level features, while progressively ambiguous cases refine higher-level emotional representations.

Early speech-related CL studies relied on heuristics, e.g., SNR-based ranking for noise-robust recognition \cite{Braun2017_Accordion} or progressively ordering training data for speaker recognition \cite{Ranjan2018_SpeakerCL}. Later work expanded to dataset-level curricula, for instance, treating acted corpora as easy and large-scale voice data on the Internet as difficult \cite{Zhou2022_AAAI_CL}, or hybrid designs combining utterance- and conversation-level difficulty \cite{Yang2021_AAAI_ConvCL}. Recent approaches adopt model-based definitions of difficulty, such as mutual information \cite{Lin2024_DeepMI} and partial information decomposition \cite{pid}.
Difficulty for human perception can be naturally explored to develop curricula. Lotfian and Busso \cite{busso} defined perception difficulty in SER as inter-annotator disagreement. They used quantitative disagreement measures such as entropy to rank samples.  

{\it In this paper, we propose CHUCKLE, a novel human perception-centered CL framework for SER that integrates score-based strategies (entropy, proportion of intended-emotion votes) (Section~\ref{subsubsec:non_rule_based}) with novel rule-based curricula (Section~\ref{subsubsec:rule_based}) that explicitly model the relationship between intended and perceived labels by jointly considering agreement strength and intended-label alignment.} 
Intuitively, assuming samples difficult for humans are likewise challenging for models, training progresses from unambiguous to ambiguous cases to improve both accuracy and efficiency. Specifically, in an acted emotion dataset, each sample has one intended emotion label and multiple perceived labels from annotators. In addition to agreement/disagreement among annotators, alignment/misalignment between intended and perceived emotions is also a strong indicator of perception difficulty. We develop curricula based on different ranking rules for agreement and alignment. Unlike prior work, CHUCKLE benchmarks multiple curricula, showing rule-based approaches outperform score-based ones. Furthermore, CHUCKLE improves training efficiency, achieving comparable performance with fewer gradient updates. Our contributions are three-fold: 
\begin{enumerate}
\item We develop a novel perception-driven CL framework that integrates rule-based and score-based curricula.
\item Our rule-based curricula consistently outperform baseline and score-based curricula. 
\item We demonstrate that CHUCKLE converges faster with better performance and fewer gradient updates. 
\end{enumerate}

\section{Dataset}
\label{ssec:subheaddataset}
We used the CREMA-D~\cite{CREMA-D} dataset, a standard audiovisual benchmark comprising 7,442 clips from 91 actors expressing 6 emotions in 12 sentences. Each clip has one intended label and multiple perceived labels (8--12 ratings) from 2,443 raters, yielding four types of labels per clip (one perceived per modality and one intended). These multiple labels make CREMA-D suitable for perception-based curriculum design, which is uncommon in SER datasets. Table~\ref{tab:matching_percentages} shows agreement between perceived and intended emotions, with audio agreement ranging from 95.7\% (neutral) to 16.4\% (sadness), underscoring the difficulty of recognizing certain emotions from speech alone.
\begin{table}[!htbp]
\centering
\caption{Matching rates (\%) of perceived vs. intended emotions}
\label{tab:matching_percentages}
\resizebox{\columnwidth}{!}{
\begin{tabular}{|c|c|c|c|c|c|c|c|}
\hline
\textbf{Modality} & \textbf{ANG} & \textbf{DIS} & \textbf{FEA} & \textbf{HAP} & \textbf{NEU} & \textbf{SAD} & \textbf{ALL} \\
\hline
Audio & 60.6 & 30.0 & 32.0 & 26.0 & 95.7 & 16.4 & 41.6 \\
\hline
Video & 65.4 & 63.4 & 51.6 & 95.6 & 91.8 & 33.4 & 64.3 \\
\hline
Multimodal & 74.7 & 74.4 & 64.8 & 94.8 & 95.7 & 32.3 & 72.2 \\
\hline
\end{tabular}
}
\end{table}

\section{Methodology}
\label{sec:pagestyle}
\subsection{Curriculum Design}
\label{ssec:curriculum_design}

The design of curricula for SER must account for the subjective and often ambiguous nature of emotional labels. In acted datasets such as CREMA-D, the agreement between intended and perceived labels shows how consistently an expression is recognized, while disagreement indicates ambiguity or possible misinterpretation. From a learning perspective, clips with strong agreement serve as clear, low-ambiguity signals that help models establish reliable low-level features, while clips with disagreement or ambiguity are more challenging samples that push models to refine and better generalize. We propose two complementary curriculum strategies:

\subsubsection{Score-Based Curricula}
\label{subsubsec:non_rule_based}

Score-based curricula use continuous scores to quantify label ambiguity. The samples are equally divided into quartiles: \texttt{Easy}, \texttt{Borderline Easy}, \texttt{Borderline Tough}, and \texttt{Tough} based on the difficulty scores.

\textbf{Intended Emotion Score:} This curriculum assigns each sample a score based on the proportion of annotators who agree with the actor’s intended emotion, where higher agreement indicates clearer and more reliable samples.
\begin{align}        
Score(intended\_emotion) = \frac{n_{\text{intended}}}{N}
\end{align}
where $n_{\text{intended}}$ is the number of annotators selecting the intended emotion and $N$ is the total number of annotators.

\textbf{Entropy Score:} This curriculum assigns scores using Shannon entropy $H$ \cite{entropy} over annotator label distributions. Higher entropy indicates greater disagreement, marking the sample as more challenging.
\begin{align}
    Score(entropy) = -\sum_{i=1}^{n} p_i \log p_i 
\end{align}
where $p_i$ denotes the fraction of annotations with emotion $i$. 
    
\subsubsection{Rule-Based Curricula}
\label{subsubsec:rule_based}

Rule-based curricula are  based on the relationships between the majority of perceived emotions and the intended emotion. They can be categorized as follows: (1) \textbf{Clear Match:} There is a clear majority among the perceived emotions, and the majority matches the intended emotion; (2) \textbf{Clear Mismatch:}  The clear majority of the perceived emotions does not match the intended emotion; (3) \textbf{Ambiguous Match:} The annotations distribute among several emotions without a dominating majority. Multiple annotations are consistent with the intended emotion. (4) \textbf{Ambiguous Mismatch:} There is no majority perceived emotion, and the intended emotion does not receive multiple annotations. Data sample counts by category are reported in Table~\ref{tab:data_distribution_rule_based}.

\vspace{-0.53em}
\begin{table}[htb]
\centering
\caption{Data Distribution over Four Categories}
\label{tab:data_distribution_rule_based}
\resizebox{\columnwidth}{!}{
\begin{tabular}{|c|c|c|c|}
\hline
\textbf{Clear Match} & \textbf{Clear Mismatch} & \textbf{Ambiguous Match} & \textbf{Ambiguous Mismatch} \\
\hline
 3,099 & 3,699 & 464 & 180 \\
\hline
\end{tabular}
}
\end{table}

To develop a curriculum from easy to hard, one should order the relative difficulty of these four categories. All orderings start with \textbf{Clear Match (1)} as \texttt{Easy}, as these samples show strong agreement that matches the intended emotion. They are unambiguous, reliable, and therefore the easiest to learn from. The differences across curricula stem from how they order the other categories (2, 3, and 4) and whether they prioritize \textit{agreement strength} or \textit{intended-label alignment}.

\textbf{Intended-Perceived Agreement 1 (1 $\rightarrow$ 2 $\rightarrow$ 3 $\rightarrow$ 4):}
This curriculum prioritizes agreement strength over alignment. After \textbf{Clear Match (1)}, \textbf{Clear Mismatch (2)} comes as \texttt{Borderline Easy} because, although the perceived label differs from the intended one, annotators strongly agree on a single alternative. \textbf{Ambiguous Match (3)} is  \texttt{Borderline Tough} because there are multiple agreements between annotators, suggesting partial ambiguity, but still one of the agreements matches the intended emotion. \textbf{Ambiguous Mismatch (4)} is \texttt{Tough}, as there are multiple agreements between annotators, but none of them match the intended label. This combined ambiguity and misalignment make the sample highly confusing.

\textbf{Intended-Perceived Agreement 2 (1 $\rightarrow$ 3 $\rightarrow$ 4 $\rightarrow$ 2):} This curriculum emphasizes alignment with the intended emotion as the key factor. \textbf{Ambiguous Match (3)} follows \textbf{Clear Match (1)}, because even with multiple agreements, at least one aligns with the intended emotion. \textbf{Ambiguous Mismatch (4)} comes next, since none of the ambiguous agreements align with the intended label, introducing both conflict and misalignment. \textbf{Clear Mismatch (2)} is placed last, the strong agreement among annotators is not consistent with the intended emotion. Such ``confidently incorrect" labels are considered most misleading, as they misalign the model to learn the wrong mapping.

\begin{figure*}[htb]
\begin{minipage}[a]{\linewidth}
  \centering
  \centerline{\includegraphics[width=0.245\textwidth]{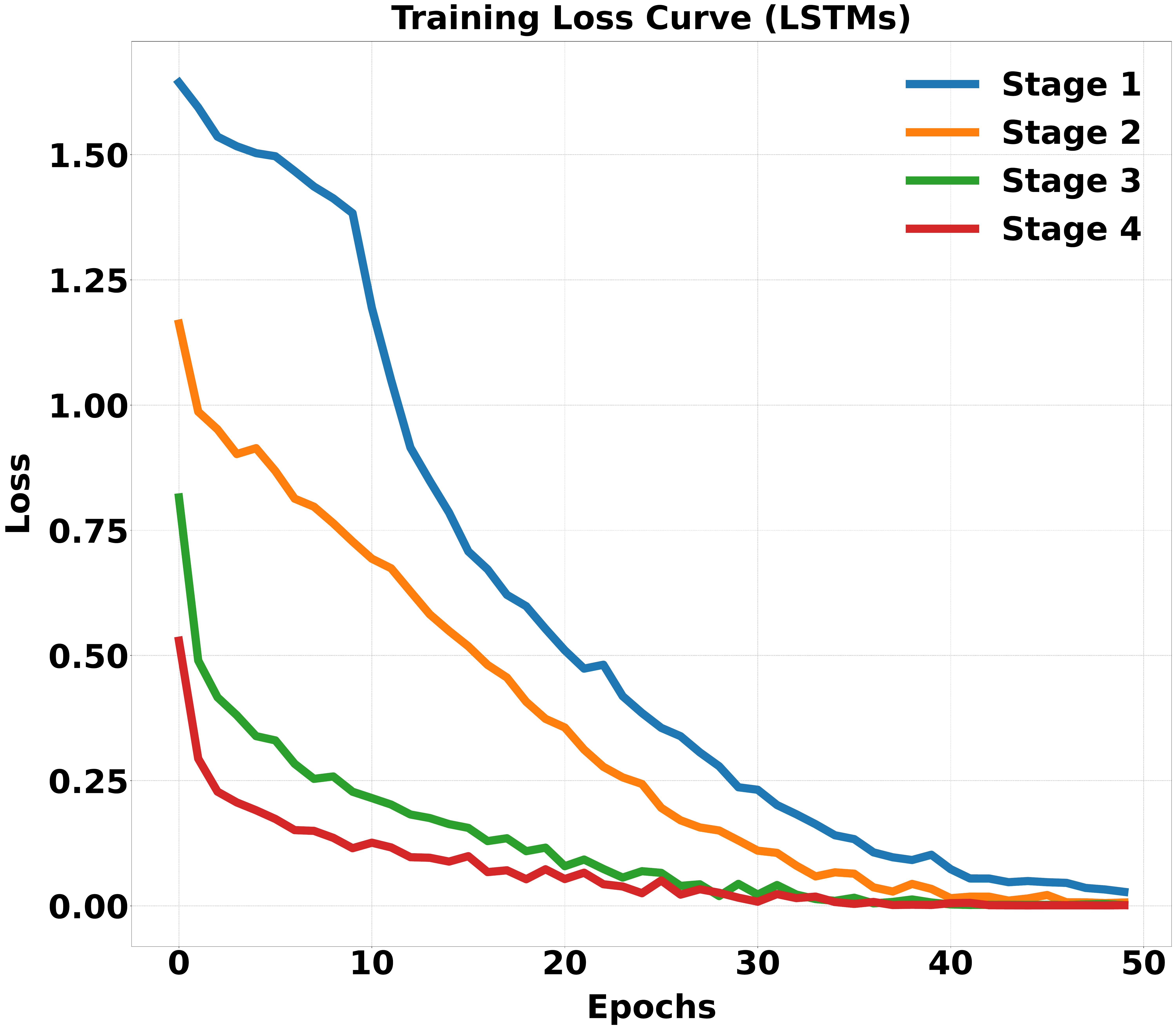}
  \includegraphics[width=0.245\textwidth]{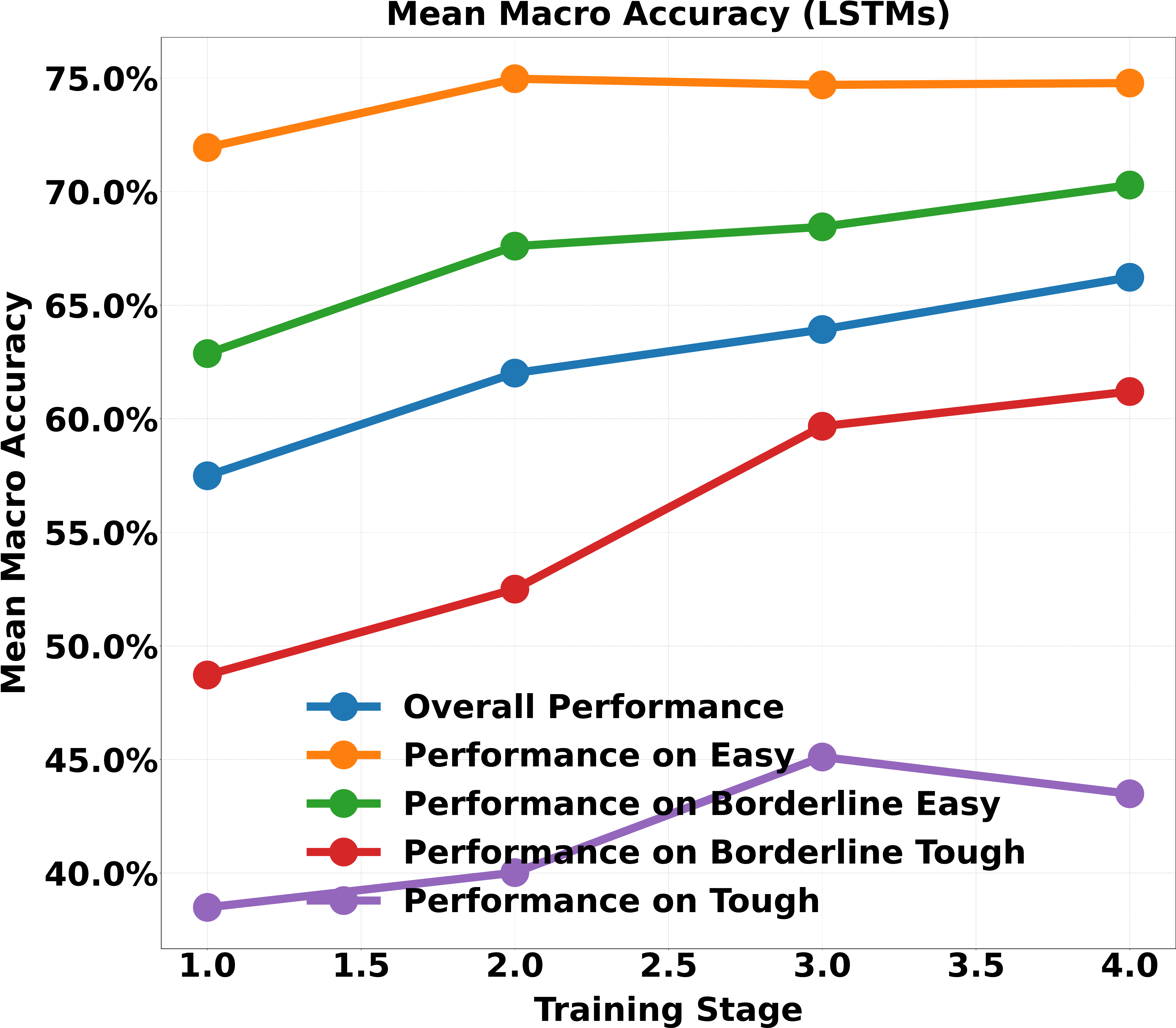}
  \includegraphics[width=0.245\textwidth]{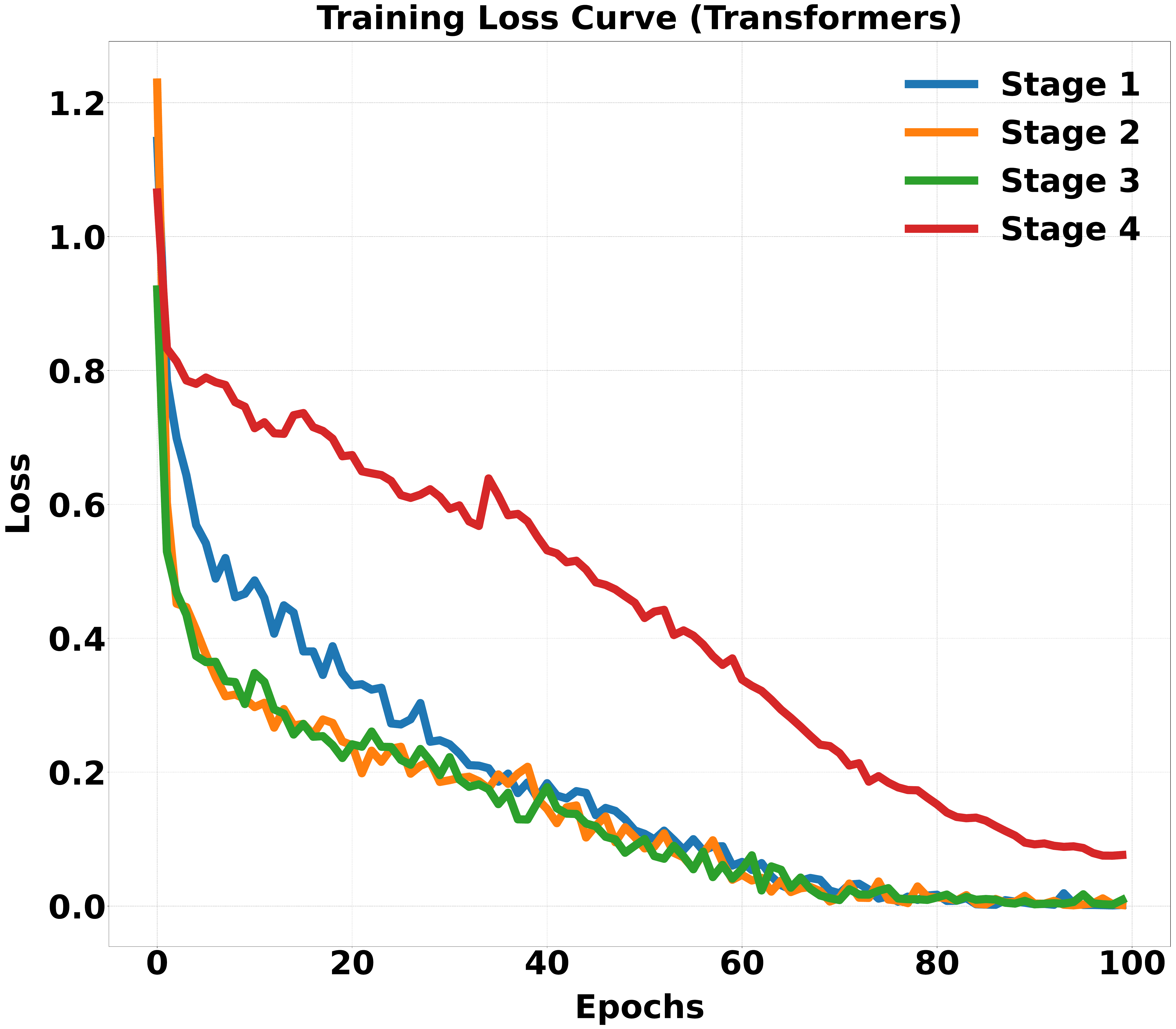}
  \includegraphics[width=0.245\textwidth]{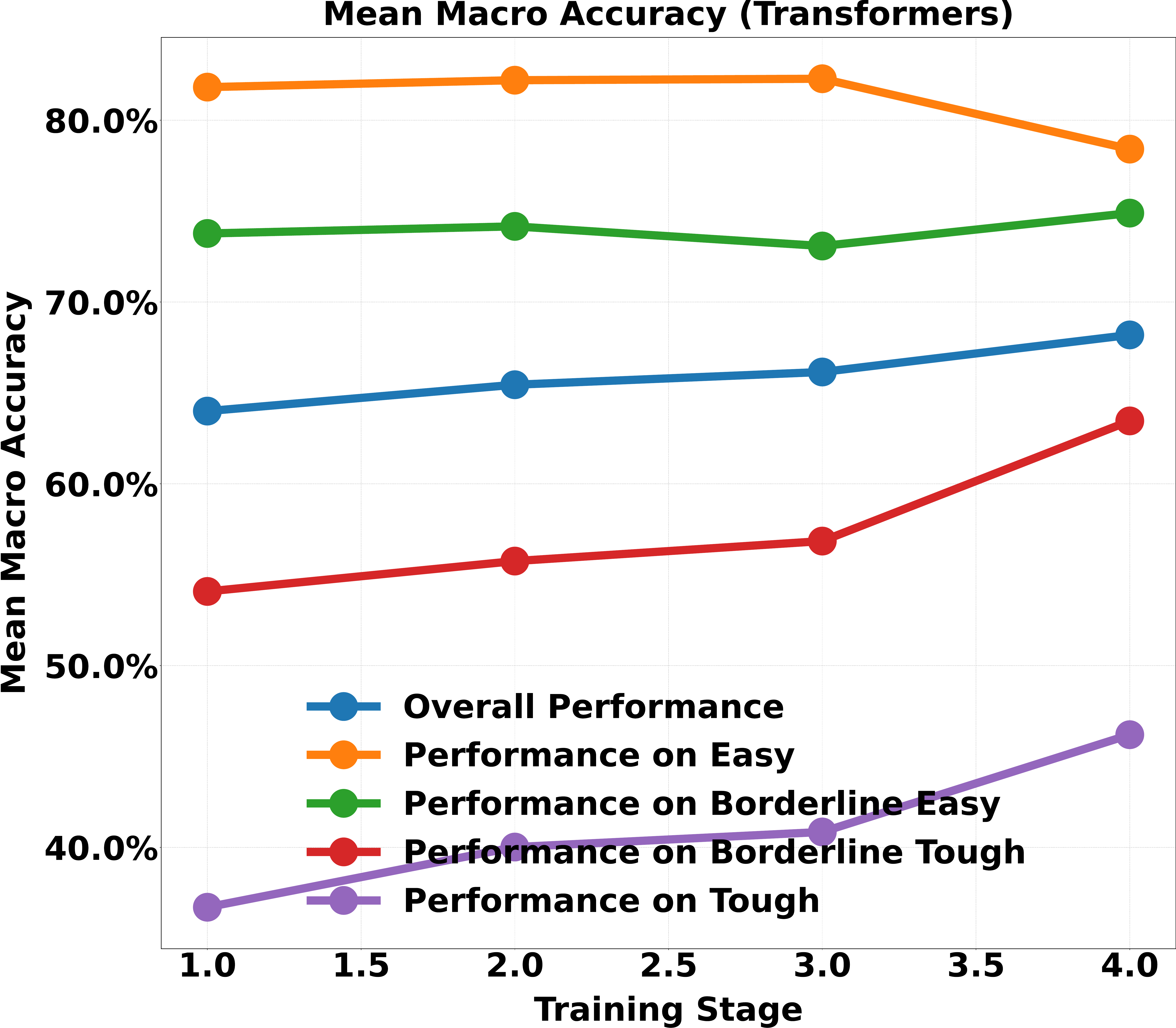}}
\end{minipage}
\caption{Model Performances for Intended-Perceived Agreement 1 on Subject Dependent Setting (Left to Right): (a) LSTM Training Loss (single trial), (b) LSTM Mean Macro Accuracy across stages (all trials), (c) Transformer Training Loss (single trial), (d) Transformer Mean Macro Accuracy across stages (all trials).}
\label{fig:model-performances}
\end{figure*}

\begin{figure*}[htb]
\begin{minipage}[a]{\linewidth}
  \centering
  \centerline{\includegraphics[width=0.245\textwidth]{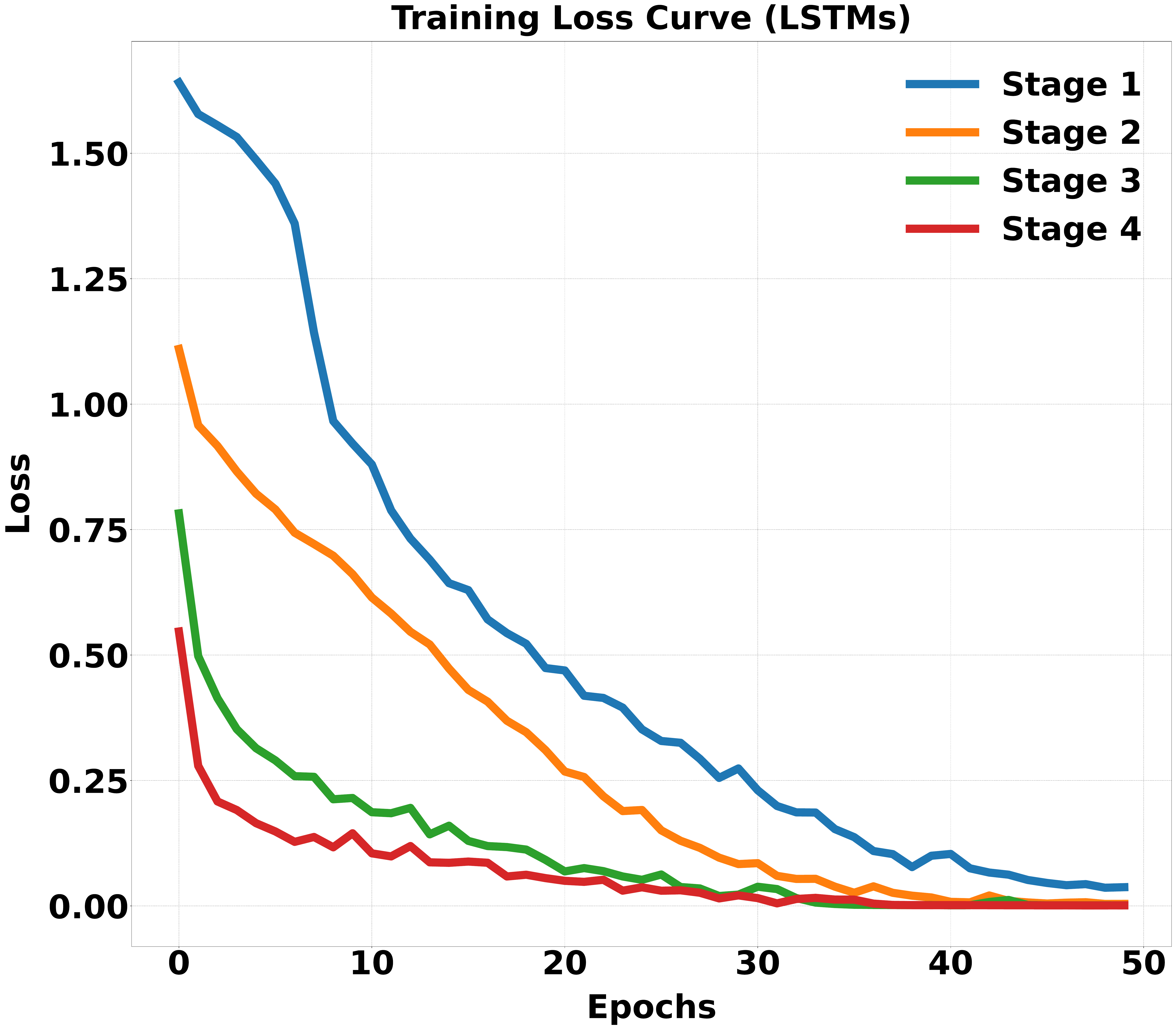}
  \includegraphics[width=0.245\textwidth]{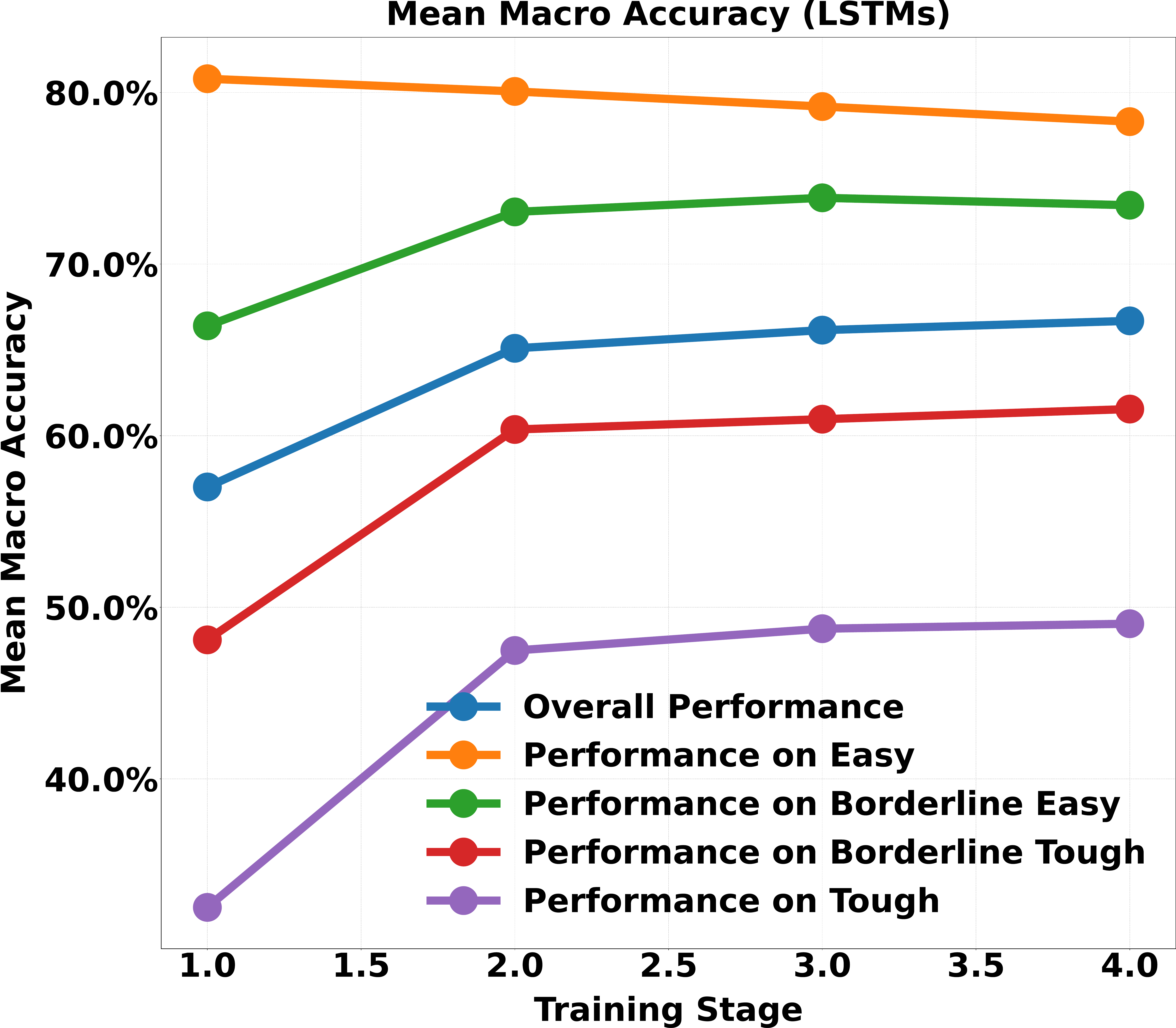}
  \includegraphics[width=0.245\textwidth]{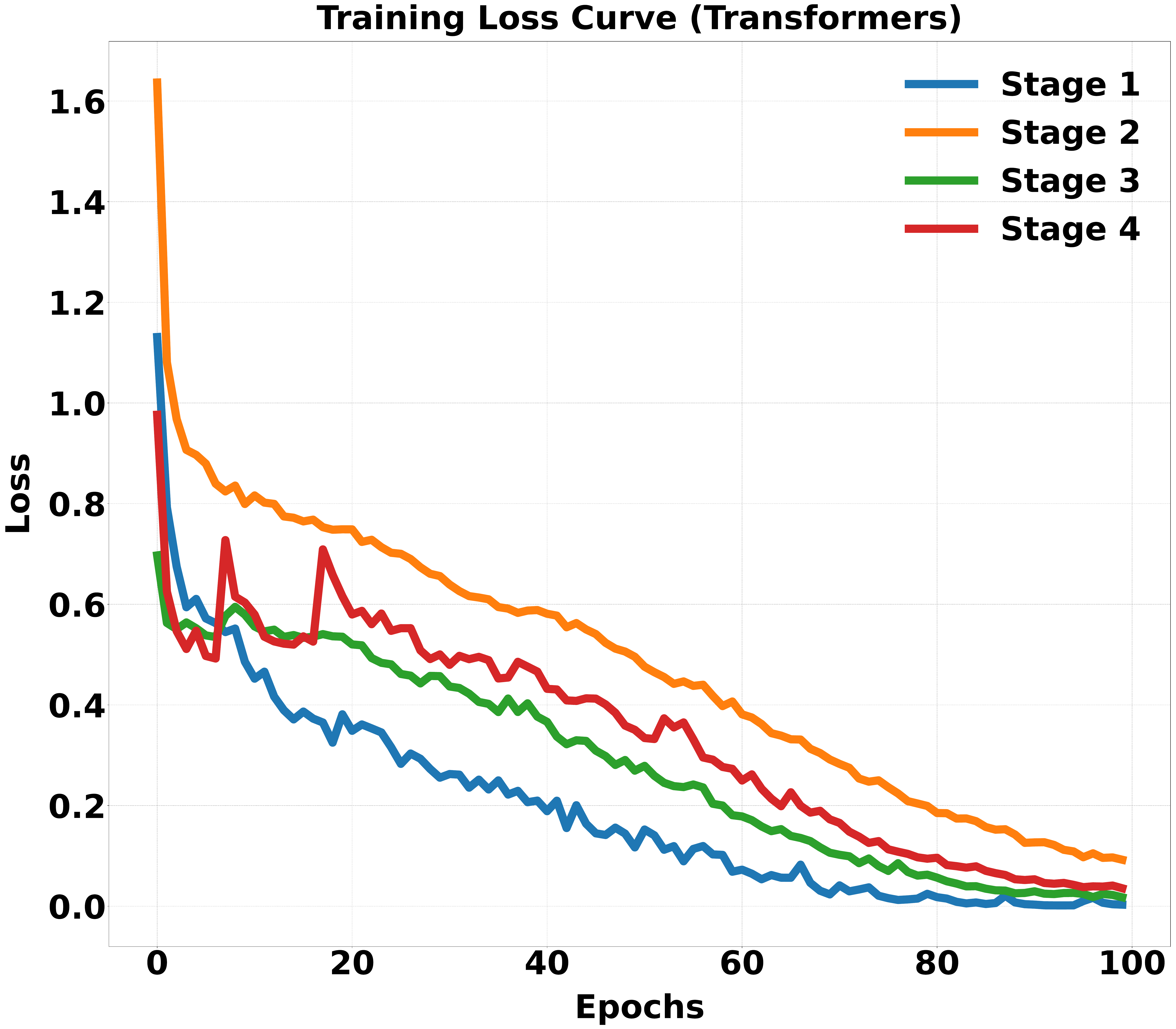}
  \includegraphics[width=0.245\textwidth]{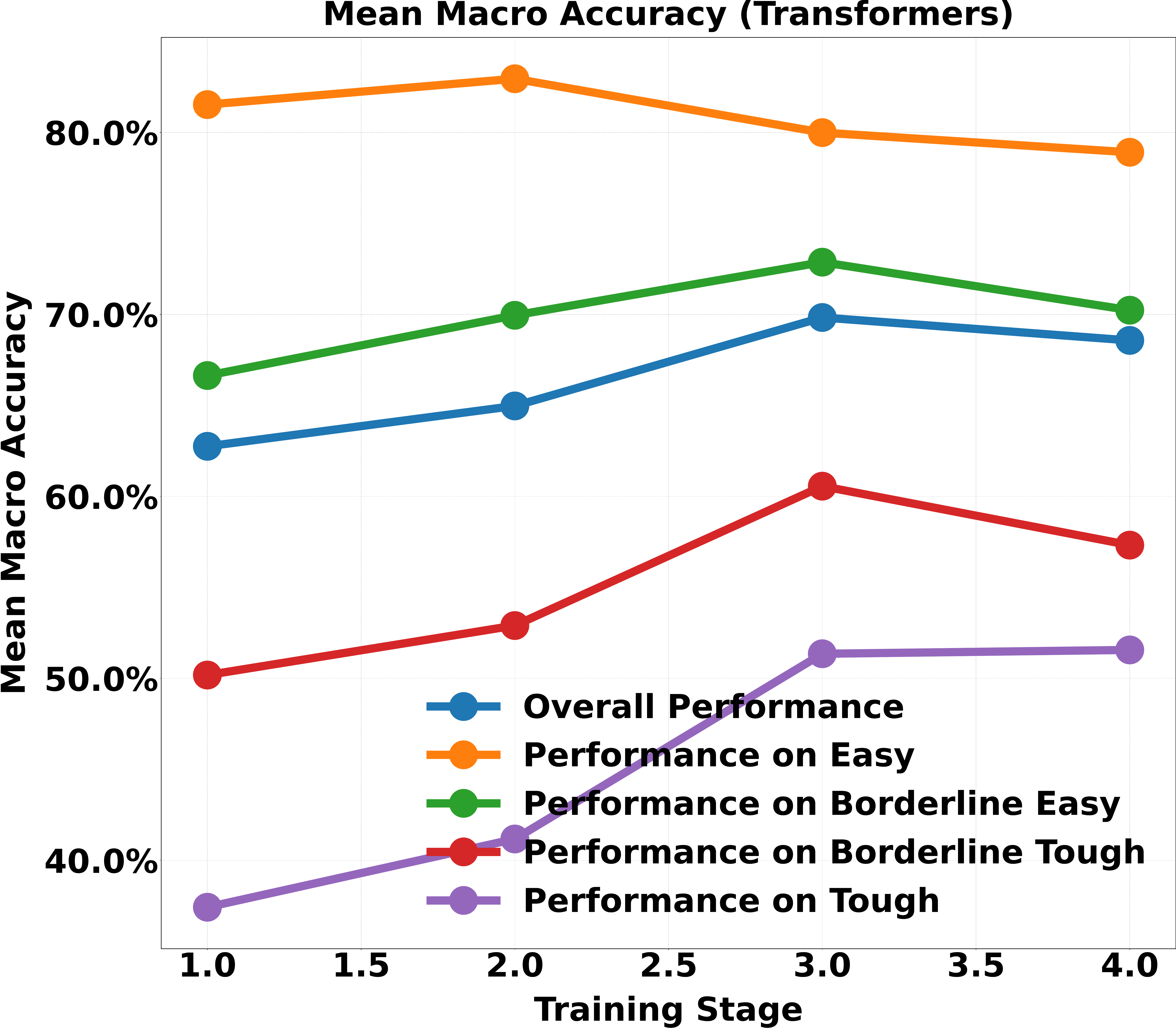}}
\end{minipage}
\caption{Model Performances for Intended-Perceived Agreement 1 on Subject Independent Setting (Left to Right): (a) LSTM Training Loss (single trial), (b) LSTM Mean Macro Accuracy across stages (all trials), (c) Transformer Training Loss (single trial), (d) Transformer Mean Macro Accuracy across stages (all trials).}
\label{fig:model-performances-SI}
\end{figure*}

\textbf{Intended-Perceived Agreement 3 (1 $\rightarrow$ 3 $\rightarrow$ 2 $\rightarrow$ 4):} This curriculum represents a compromise between alignment and agreement strength. Similar to the previous ordering, \textbf{Ambiguous Match (3)} follows \textbf{Clear Match (1)}. \textbf{Clear Mismatch (2)} comes before \textbf{Ambiguous Mismatch (4)}, because even though the majority perceived label disagrees with the intended one, the strong agreement provides a consistent (albeit incorrect) signal, which may be easier to model than \textbf{Ambiguous Mismatch (4)}, where both ambiguity and misalignment occur simultaneously. 

\subsection{Training Strategy and Computational Cost Analysis}
\label{ssec:cost_computation}

Training starts with samples in the \texttt{Easy} bin. The samples in the other bins are sequentially added in the following stages, enabling gradual learning while minimizing the risk of catastrophic forgetting.  This design reduces computation with fewer gradient updates to reach the target performance. Assuming one update per batch, total updates for non-curriculum training are:
\begin{align}
U_{\text{non-cl}} = \text{Batches per Epoch} \times \text{Total Epochs} = \left\lceil \frac{N}{B} \right\rceil \times E_{\text{total}}
\end{align} 
where \( N \) represents the total number of training samples, \( B \) is the batch size, and \( E_{\text{total}} \) is the total number of epochs. 

In $K$-stage curriculum learning, the dataset used in each stage  increases \( S_1 \subset S_2 \subset \dots \subset S_K = S_{\text{full}} \). The total updates are:
\begin{align} 
U_{\text{cl}} = \sum_{i=1}^{K} \left( \left\lceil \frac{|S_i|}{B} \right\rceil \times E_i \right)
\end{align}
where \( |S_i| \) denotes the size of the subset \( S_i \),  \( B \) is the batch size, and \( E_i \) the number of epochs assigned to the stage  \( i \).

\section{Experimental Evaluation}
\label{sec:experiment_eval}

\begin{figure*}[htb]
\begin{minipage}[a]{\linewidth}
  \centering
  \centerline{\includegraphics[width=0.45\textwidth]{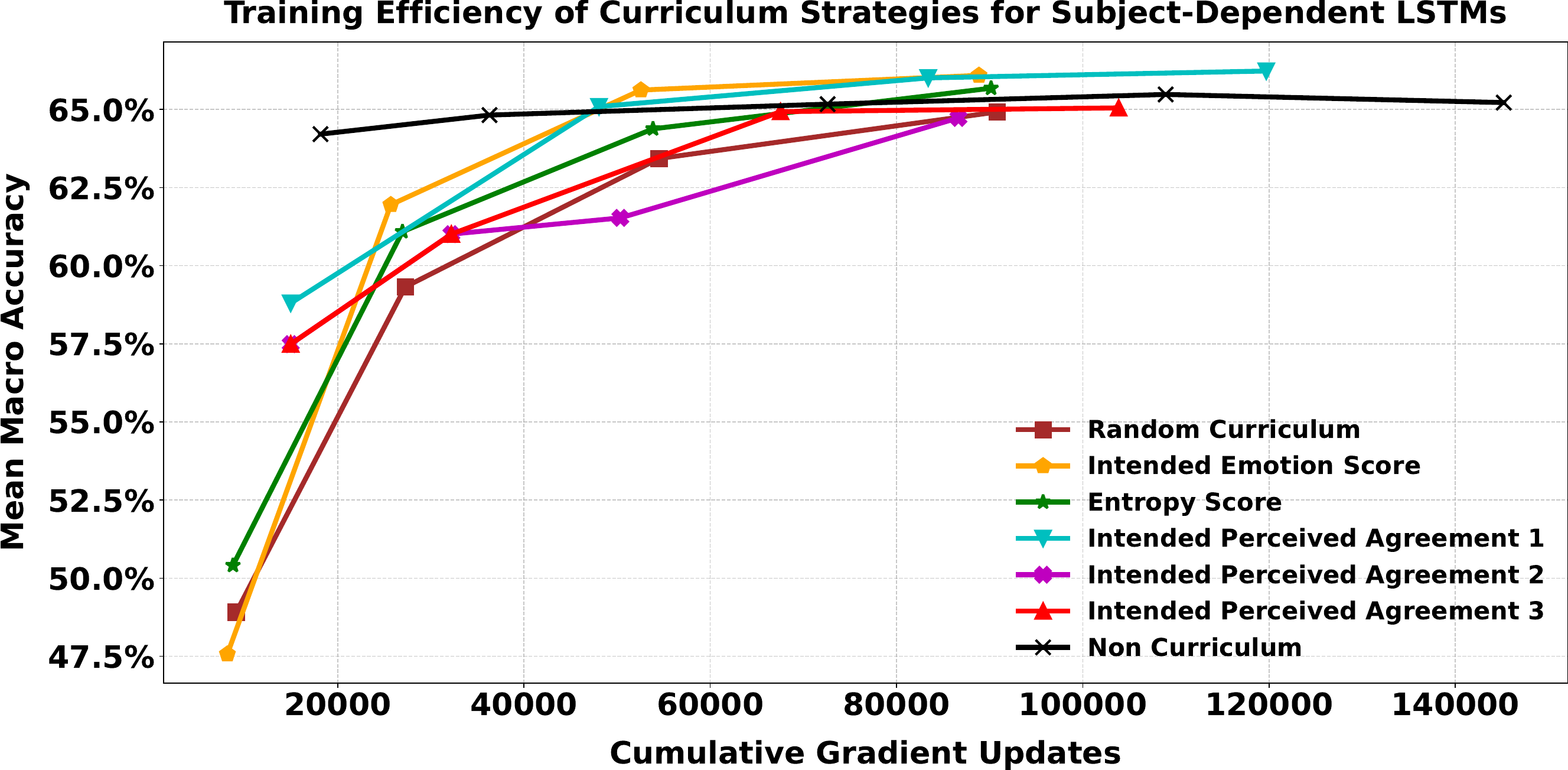}
  \includegraphics[width=0.45\textwidth]{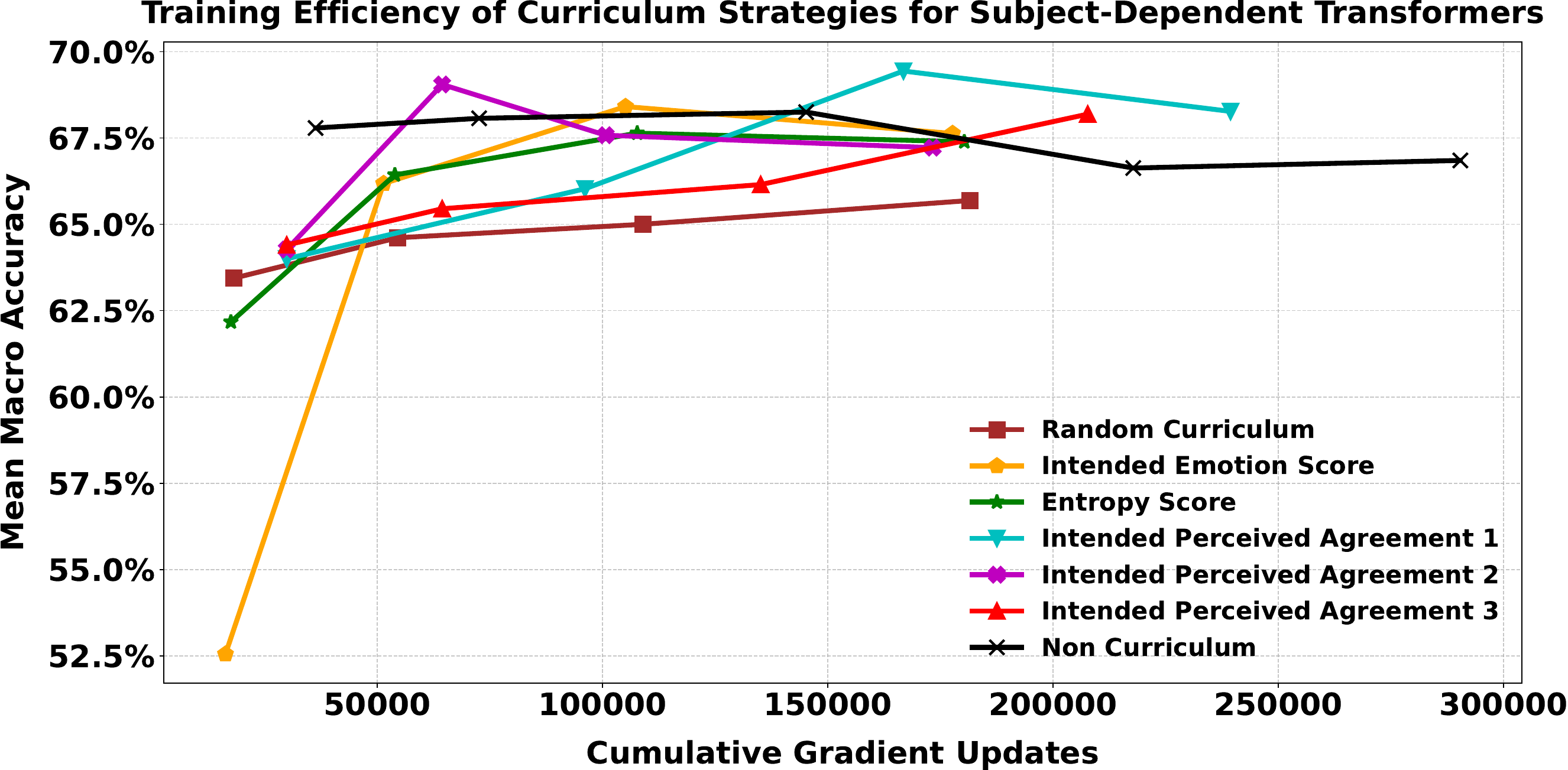}}
\end{minipage}
\caption{Training Cost  Comparisons (Gradient Updates) for Subject-Dependent (Left to Right): (a) LSTMs, (b) Transformers.}
\label{fig:cost-compare}
\end{figure*}

\begin{figure*}[htb]
\begin{minipage}[a]{\linewidth}
  \centering
  \centerline{\includegraphics[width=0.45\textwidth]{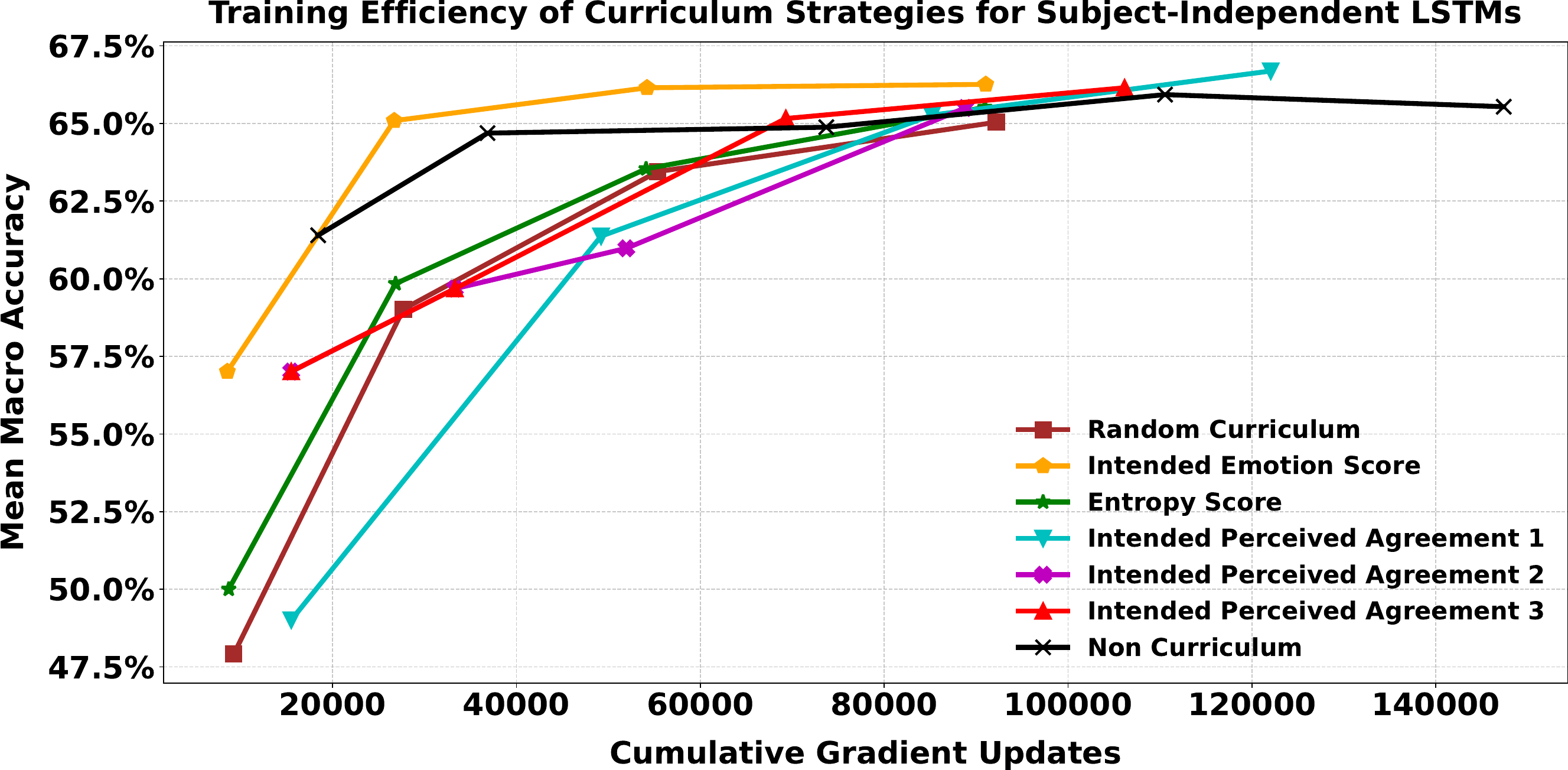}
  \includegraphics[width=0.45\textwidth]{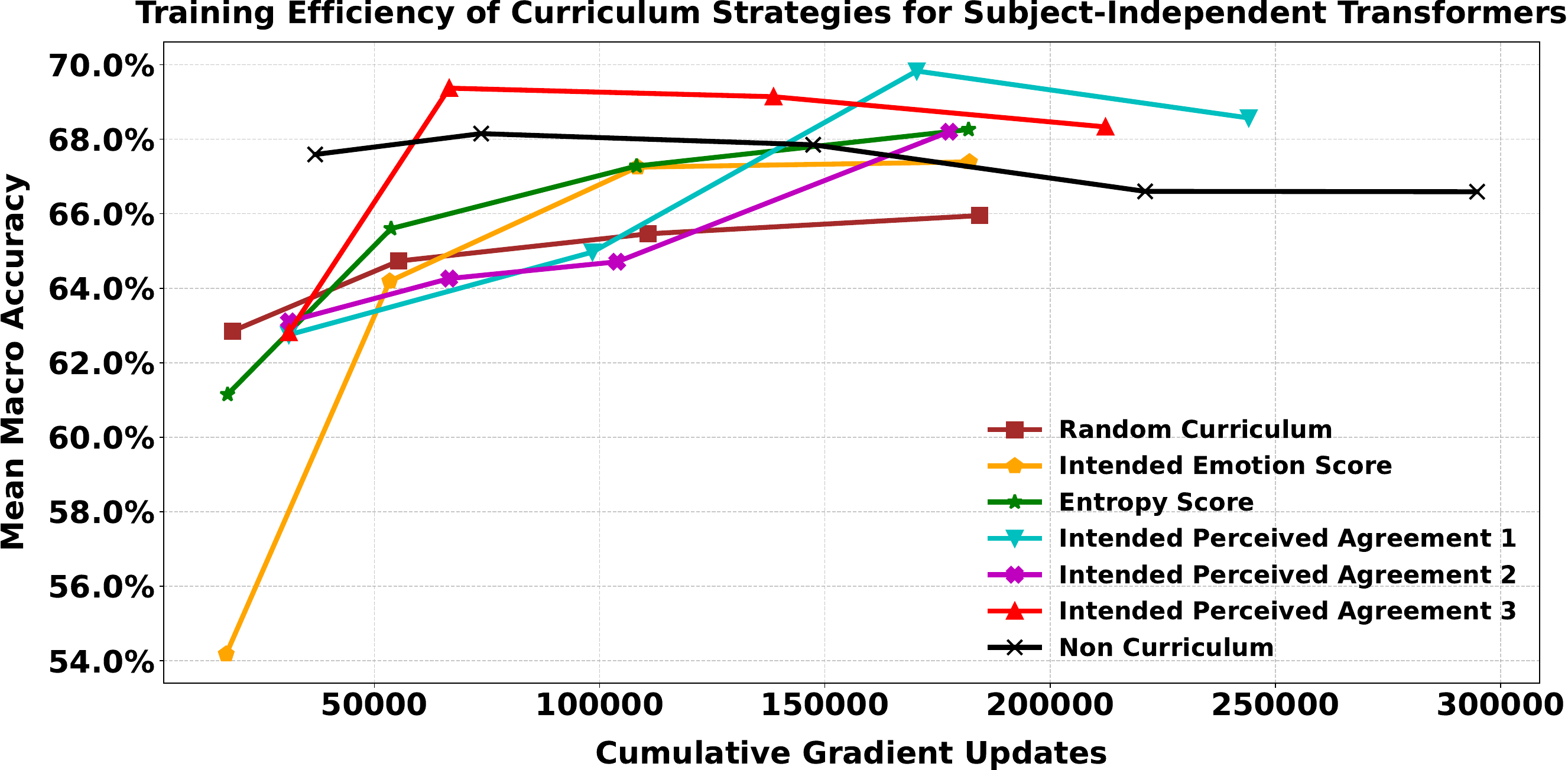}}
\end{minipage}
\caption{Training Cost  Comparisons (Gradient Updates) for Subject-Independent (Left to Right): (a) LSTMs, (b) Transformers.}
\label{fig:cost-compare-SI}
\end{figure*}

Neural networks typically require large datasets and extensive training, but limited training clips and computing resources constrained our work. To address this challenge, we used pre-extracted HuBERT representations~\cite{hsu2021hubert} instead of raw audio, enabling more efficient learning.

\vspace{-0.5em}
\begin{table}[htb]
\centering
\caption{Model Training Details}
\resizebox{\columnwidth}{!}{
\begin{tabular}{p{0.33\linewidth}p{0.35\linewidth}p{0.35\linewidth}}
\hline
\textbf{Setting} & \textbf{LSTM} & \textbf{Transformer} \\
\hline
\textbf{Architecture} 
& 2-layer BiLSTM\cite{lstm,bilstm} (128-dim) 
& 2-layer Transformer\cite{transformer} (4 heads, 128-dim) \\
\hline
\textbf{Epochs} 
& 200 (50 per stage) 
& 400 (100 per stage) \\
\hline
\textbf{Optimizer} & \multicolumn{2}{l}{Adam~\cite{ADAM}} \\
\textbf{LR Scheduler} & \multicolumn{2}{p{0.70\linewidth}}{CosineAnnealingLR~\cite{lrscheduler} (decay $5 \times 10^{-4}$ to $5 \times 10^{-5}$, reset per bin)} \\
\textbf{GPU} & \multicolumn{2}{l}{RTX A6000} \\
\textbf{Trials \& Validation} & \multicolumn{2}{p{0.70\linewidth}}{10 trials per setting, evaluated on mean macro accuracy and validated with a paired one-sided $t$-test ($p < 0.05$)} \\
\hline
\end{tabular}
}
\label{tab:model_details}
\end{table}

We used self-supervised speech representations from the HuBERT model, specifically the HuBERT-Xlarge variant available through the Hugging Face Transformers library~\cite{wolf2020transformers}. All audio samples were resampled to 16 kHz and processed with the HuBERT feature extractor. We extracted frame-level representations from the model's final hidden layer without performing task-specific fine-tuning. The HuBERT-Xlarge model generated 1280-dimensional embeddings per frame, resulting in variable-length sequences based on the duration of the audio. 

We trained in three settings: non-curriculum (using the full dataset for each epoch), random curriculum (randomly dividing the dataset into four equal sequential parts), and curriculum learning (progressively increasing difficulty using curricula in Sec.~\ref{ssec:curriculum_design}).  Table~\ref{tab:model_details} summarizes the experimental setup details.

To evaluate model performance under various generalization conditions, we used both subject-dependent and subject-independent settings. In the subject-dependent setting, clips from each subject and each intended emotion were split 80\% for training and 20\% for testing. This approach allowed samples from the same speaker to appear in both sets, helping the model learn speaker-specific traits. In contrast, the subject-independent setting strictly separated speakers, with approximately 80\% of subjects used for training and 20\% for testing. This assessment measures the model's ability to generalize to unseen speakers.

\vspace{-0.5em}
\begin{table}[htbp]
\centering
\caption{Mean Macro Accuracy (± Std.) — Subject Dependent (* denotes statistical significance at $p<0.05$)}
\label{tab:results_subjectdependent}
\resizebox{\columnwidth}{!}{
\begin{tabular}{p{0.52\linewidth}p{0.24\linewidth}p{0.24\linewidth}}
\hline
Curriculum Design
& LSTM 
& Transformer \\

\hline

Non Curriculum & 0.6522($\pm$0.015) & 0.6685($\pm$0.012) \\
Random Curriculum & 0.6492($\pm$0.011) & 0.6569($\pm$0.015) \\
\hline

Intended Emotion Score~\cite{busso} & 0.6609($\pm$0.009) & 0.6763($\pm$0.017) \\
Entropy Score~\cite{busso} & 0.6568($\pm$0.011) & 0.6739($\pm$0.009) \\
\hline

Intended Perceived Agreement 1 & \textbf{0.6623($\pm$0.010)$^{*}$} & \textbf{0.6827($\pm$0.012)$^{*}$} \\
Intended Perceived Agreement 2 & 0.6473($\pm$0.014) & 0.6722($\pm$0.023) \\
Intended Perceived Agreement 3 & 0.6505($\pm$0.012) & 0.6819($\pm$0.025)$^{*}$ \\
\hline

\end{tabular}
}
\end{table}

\vspace{-1.45em}
\begin{table}[htbp]
\centering
\caption{Mean Macro Accuracy (± Std.) — Subject Independent (* denotes statistical significance at $p<0.05$)}
\label{tab:results_subjectindependent}
\resizebox{\columnwidth}{!}{
\begin{tabular}{p{0.52\linewidth}p{0.24\linewidth}p{0.24\linewidth}}
\hline

Curriculum Design
& LSTM
& Transformer \\

\hline

Non Curriculum & 0.6554($\pm$0.012) & 0.6659($\pm$0.016) \\
Random Curriculum & 0.6504($\pm$0.018) & 0.6595($\pm$0.023) \\
\hline

Intended Emotion Score~\cite{busso} & 0.6626($\pm$0.009) & 0.6739($\pm$0.014) \\
Entropy Score~\cite{busso} & 0.6552($\pm$0.012) & 0.6826($\pm$0.009)$^{*}$ \\
\hline

Intended Perceived Agreement 1 & \textbf{0.6669($\pm$0.008)$^{*}$} & \textbf{0.6857($\pm$0.014)$^{*}$} \\
Intended Perceived Agreement 2 & 0.6550($\pm$0.014) & 0.6820($\pm$0.013)$^{*}$ \\
Intended Perceived Agreement 3 & 0.6615($\pm$0.012) & 0.6833($\pm$0.021)$^{*}$ \\
\hline

\end{tabular}
}
\end{table}

\textbf{LSTMs}: Although the absolute gains are modest, they are consistent across architectures and settings, and are achieved without increasing model complexity. In the subject-dependent setting (Table~\ref{tab:results_subjectdependent}), most curricula  performed similar or better than the non-curriculum baseline, except for the random curriculum, which showed a drop in accuracy. Intended-Perceived Agreement 1 achieved the best performance with a significant improvement, indicating that LSTMs benefit from perception-driven curricula. A similar pattern can be observed in the subject-independent setting (Table~\ref{tab:results_subjectindependent}), where Intended-Perceived Agreement 1 again yielded the highest accuracy and was statistically significant. Overall performance slightly exceeded the subject-dependent results, suggesting that perception-based CL is beneficial, even with unseen speakers. Figures\ref{fig:model-performances}(a) and \ref{fig:model-performances-SI}(a) show that CL reduced initial errors and sped up convergence by leveraging knowledge from earlier stages. Figures\ref{fig:model-performances}(b) and \ref{fig:model-performances-SI}(b) show steady improvements as the model moved from easier to more difficult samples, with the most significant gains between the \texttt{Easy} and \texttt{Borderline Easy} bins, which contained most of the data.

\textbf{Transformer Models:} Transformers outperformed LSTMs in baseline accuracy, aligning with their capability to model long-range dependencies effectively. In the subject-dependent scenario (Table~\ref{tab:results_subjectdependent}), most curricula showed improvements, with the Intended-Perceived Agreement 1 approach achieving the best statistically significant result, while the random curriculum slightly worsened performance. This suggests that Transformers are less resilient to poorly structured curricula. In the subject-independent setting (Table~\ref{tab:results_subjectindependent}), curriculum learning had a more noticeable impact, with several methods, particularly Intended-Perceived Agreements 1, 2, 3, and an entropy-based curriculum, achieving significant improvements. Intended-Perceived Agreement 1 outperformed the others, indicating effective generalization across speakers. Training loss (Figure~\ref{fig:model-performances}(c) and ~\ref{fig:model-performances-SI}(c)) decreased across stages, with a spike at Stage 2 from the larger and more diverse \texttt{Borderline Easy} subset. Although the transformers were more affected by distribution changes, they stabilized over time, and the macro accuracy (Figure~\ref{fig:model-performances} (d) and ~\ref{fig:model-performances-SI}(d)) improved consistently across stages.

\textbf{Cost Analysis:} Figures~\ref{fig:cost-compare} and \ref{fig:cost-compare-SI} show that CL improves training efficiency for both LSTMs and Transformers in subject-dependent (SD) and subject-independent (SI) settings. For LSTMs, rule-based curricula improve relative accuracy by 0.7\% to 1.6\% in SD and 0.9\% to 1.8\% in SI. Intended-Perceived Agreement 1 achieves the highest mean accuracy, with 1.6\% relative improvement in SD and 1.8\% in SI, while using 17\% fewer updates than the baseline likely because strong annotator agreement offers a stable learning signal, even when misaligned with intended labels. Intended-Perceived Agreement 2 reduces training costs by nearly 40\% with comparable performance. Transformers see larger gains, with relative mean accuracy improvements of 0.8\% to 2.1\% in SD and 1.2\% to 3.0\% in SI. Again, Intended-Perceived Agreement 1 shows the highest relative improvement, with about 2.1\% in SD and 3.0\% in SI, alongside a reduction of about 17\% in updates. Similar to LSTMs, Intended-Perceived Agreement 2 reduces updates by nearly 40\% in both settings with minimal impact on accuracy.

\section{Conclusion and Future Work}
\label{sec:conclusion}
We demonstrated that annotator agreement and alignment in crowd-sourced datasets serve as a high-value input for CL in SER. Rule-based curricula outperformed non-curriculum and score-based curricula in both accuracy and efficiency. LSTMs showed relative gains of 1.6\% in SD and 1.8\% in SI with rule-based curricula, while Transformers achieved relative gains of 2.1\% in SD and 3.0\% in SI. These curricula also reduced training costs by requiring nearly 40\% fewer gradient updates with comparable performance. Overall, CHUCKLE enhances performance and efficiency, especially in cross-speaker scenarios. Since CHUCKLE operates at the sample-ordering level, it is model-agnostic and can be applied to different architectures whenever intended and perceived labels are available. Future work will explore these curricula for multimodal emotion recognition across various datasets and models.

\section{Generative AI Use Disclosure}

Generative AI tools were utilized to improve the clarity and readability of certain sections of this manuscript. This included minor language editing and suggestions for phrasing. These tools were not used to generate scientific ideas, experimental designs, results, or conclusions. All technical content, analysis, and interpretations were developed by the authors, who assume full responsibility for the accuracy and integrity of the work.

\bibliographystyle{IEEEtran}
\bibliography{mybib}

@inproceedings{cl_bengio,
  author    = {Y. Bengio and J. Louradour and R. Collobert and J. Weston},
  title     = {Curriculum learning},
  booktitle = {Proceedings of the 26th Annual International Conference on Machine Learning},
  year      = {2009},
  pages     = {41-48},
  address   = {Montreal, QC, Canada},
  doi       = {10.1145/1553374.1553380},
  url       = {https://doi.org/10.1145/1553374.1553380}
}

@inproceedings{transformer,
 author  = {A. Vaswani and N. Shazeer and N. Parmar and J. Uszkoreit and L. Jones and A. N. Gomez and L. Kaiser and I. Polosukhin},
 booktitle = {Advances in Neural Information Processing Systems},
 pages = {},
 year = {2017},
 volume = {30},
 title = {Attention is All you Need},
 url ={https://proceedings.neurips.cc/paper_files/paper/2017/file/3f5ee243547dee91fbd053c1c4a845aa-Paper.pdf},
}

@inproceedings{ADAM,
  author       = {D. P. Kingma and J. Ba},
  title        = {Adam: {A} Method for Stochastic Optimization},
  booktitle    = {3rd International Conference on Learning Representations, {ICLR} 2015,
                  San Diego, CA, USA, May 7-9, 2015, Conference Track Proceedings},
  year         = {2015},
  url          = {http://arxiv.org/abs/1412.6980},
  bibsource    = {dblp computer science bibliography, https://dblp.org}
}

@inproceedings{lrscheduler,
  author       = {I. Loshchilov and F. Hutter},
  title        = {{SGDR:} Stochastic Gradient Descent with Warm Restarts},
  booktitle    = {5th International Conference on Learning Representations, {ICLR} 2017,
                  Toulon, France, April 24-26, 2017, Conference Track Proceedings},
  year         = {2017},
  url          = {https://openreview.net/forum?id=Skq89Scxx},
  bibsource    = {dblp computer science bibliography, https://dblp.org}
}

@INPROCEEDINGS{Braun2017_Accordion,
  author={S. Braun and D. Neil and S.C. Liu},
  booktitle={2017 25th European Signal Processing Conference (EUSIPCO)}, 
  title={A curriculum learning method for improved noise robustness in automatic speech recognition}, 
  year={2017},
  volume={},
  number={},
  pages={548-552},
  doi={10.23919/EUSIPCO.2017.8081267}}

@ARTICLE{CREMA-D,
  author={H. Cao and D. G. Cooper and M. K. Keutmann and R. C. Gur and A. Nenkova and R. Verma},
  journal={IEEE Transactions on Affective Computing}, 
  title={CREMA-D: Crowd-Sourced Emotional Multimodal Actors Dataset}, 
  year={2014},
  volume={5},
  number={4},
  pages={377-390},
  doi={10.1109/TAFFC.2014.2336244}}

@ARTICLE{entropy,
  author={C.E. Shannon},
  journal={The Bell System Technical Journal}, 
  title={A mathematical theory of communication}, 
  year={1948},
  volume={27},
  number={3},
  pages={379-423},
  doi={10.1002/j.1538-7305.1948.tb01338.x}}

@ARTICLE{bilstm,
  author={M. Schuster and K.K. Paliwal},
  journal={IEEE Transactions on Signal Processing}, 
  title={Bidirectional recurrent neural networks}, 
  year={1997},
  volume={45},
  number={11},
  pages={2673-2681},
  doi={10.1109/78.650093}}

@article{schuller2018speech, 
author = {Schuller, B. W.}, 
title = {Speech emotion recognition: two decades in a nutshell, benchmarks, and ongoing trends}, 
year = {2018}, 
issue_date = {May 2018}, 
publisher = {Association for Computing Machinery}, 
address = {New York, NY, USA}, 
volume = {61}, 
number = {5}, 
issn = {0001-0782}, 
url = {https://doi.org/10.1145/3129340}, 
doi = {10.1145/3129340}, 
journal = {Communications of the ACM}, 
month = apr, 
pages = {90–99}, 
numpages = {10} }

@article{lian2023survey,
  author={H. Lian and C. Lu and S. Li and Y. Zhao and C. Tang and Y. Zong},
  title={A Survey of Deep Learning-Based Multimodal Emotion Recognition: Speech, Text, and Face},
  journal={Entropy},
  volume={25},
  number={10},
  pages={1440},
  year={2023},
  publisher={MDPI},
  doi={10.3390/e25101440}
}

@article{CANAL2022593,
title = {A survey on facial emotion recognition techniques: A state-of-the-art literature review},
journal = {Information Sciences},
volume = {582},
pages = {593-617},
year = {2022},
issn = {0020-0255},
doi = {https://doi.org/10.1016/j.ins.2021.10.005},
url = {https://www.sciencedirect.com/science/article/pii/S0020025521010136},
author = {F. Z. Canal and T. R. Müller and J. C. Matias and G. G. Scotton and A. R. de Sa Junior and E. Pozzebon and A. C. Sobieranski},
}

@ARTICLE{noroozi,
  author={F. Noroozi  and C.A. Corneanu and D. Kamińska and T. Sapiński and S. Escalera and G. Anbarjafari},
  journal={IEEE Transactions on Affective Computing}, 
  title={Survey on Emotional Body Gesture Recognition}, 
  year={2021},
  volume={12},
  number={2},
  pages={505-523},
  doi={10.1109/TAFFC.2018.2874986}
}

@article{gandi,
author = {R. Gandi and A.Geetha and B.R. Reddy},
year = {2025},
month = {01},
pages = {},
title = {Comprehensive Survey on Recognition of Emotions from Body Gestures},
volume = {5},
journal = {Journal of Informatics Education and Research},
doi = {10.52783/jier.v5i1.2029}
}

@Article{wenqian,
AUTHOR = {W. Lin and C. Li},
TITLE = {Review of Studies on Emotion Recognition and Judgment Based on Physiological Signals},
JOURNAL = {Applied Sciences},
VOLUME = {13},
YEAR = {2023},
NUMBER = {4},
ARTICLE-NUMBER = {2573},
URL = {https://www.mdpi.com/2076-3417/13/4/2573},
ISSN = {2076-3417},
DOI = {10.3390/app13042573}
}

@article{wang,
author = {Y. Wang and B. Zhang and L. Di},
title = {Research Progress of EEG-Based Emotion Recognition: A Survey},
year = {2024},
issue_date = {November 2024},
publisher = {Association for Computing Machinery},
address = {New York, NY, USA},
volume = {56},
number = {11},
issn = {0360-0300},
url = {https://doi.org/10.1145/3666002},
doi = {10.1145/3666002},
journal = {ACM Comput. Surv.},
month = jul,
articleno = {288},
numpages = {49}
}

@article{GEORGE2024127015,
title = {A review on speech emotion recognition: A survey, recent advances, challenges, and the influence of noise},
journal = {Neurocomputing},
volume = {568},
pages = {127015},
year = {2024},
issn = {0925-2312},
doi = {https://doi.org/10.1016/j.neucom.2023.127015},
url = {https://www.sciencedirect.com/science/article/pii/S0925231223011384},
author = {S.M. George and P.M. Ilyas}
}

@ARTICLE{busso,
  author={R. Lotfian and C. Busso},
  journal={IEEE/ACM Transactions on Audio, Speech, and Language Processing}, 
  title={Curriculum Learning for Speech Emotion Recognition From Crowdsourced Labels}, 
  year={2019},
  volume={27},
  number={4},
  pages={815-826},
  doi={10.1109/TASLP.2019.2898816}}

@article{Lin2024_DeepMI,
author = {W.C. Lin and K. Sridhar and C. Busso},
title = {An Interpretable Deep Mutual Information Curriculum Metric for a Robust and Generalized Speech Emotion Recognition System},
year = {2024},
issue_date = {2024},
publisher = {IEEE Press},
volume = {32},
issn = {2329-9290},
url = {https://doi.org/10.1109/TASLP.2024.3507562},
doi = {10.1109/TASLP.2024.3507562},
journal = {IEEE/ACM Trans. Audio, Speech and Lang. Proc.},
month = nov,
pages = {5117–5130},
numpages = {14}
}

@article{Ranjan2018_SpeakerCL,
author = {S. Ranjan and J.H.L. Hansen},
title = {Curriculum Learning Based Approaches for Noise Robust Speaker Recognition},
year = {2018},
issue_date = {January 2018},
publisher = {IEEE Press},
volume = {26},
number = {1},
issn = {2329-9290},
url = {https://doi.org/10.1109/TASLP.2017.2765832},
doi = {10.1109/TASLP.2017.2765832},
journal = {IEEE/ACM Trans. Audio, Speech and Lang. Proc.},
month = jan,
pages = {197–210},
numpages = {14}
}

@article{Yang2021_AAAI_ConvCL, 
title={Hybrid Curriculum Learning for Emotion Recognition in Conversation}, 
volume={36}, 
url={https://ojs.aaai.org/index.php/AAAI/article/view/21413}, 
DOI={10.1609/aaai.v36i10.21413}, 
number={10}, 
journal={Proceedings of the AAAI Conference on Artificial Intelligence}, 
author={L. Yang and Y. Shen and Y. Mao and L. Cai}, 
year={2022}, 
month={Jun.}, 
pages={11595-11603} }

@article{Zhou2022_AAAI_CL, 
title={Inferring Emotion from Large-scale Internet Voice Data: A Semi-supervised Curriculum Augmentation based Deep Learning Approach}, volume={35}, 
url={https://ojs.aaai.org/index.php/AAAI/article/view/16753}, 
DOI={10.1609/aaai.v35i7.16753}, 
number={7}, 
journal={Proceedings of the AAAI Conference on Artificial Intelligence}, 
author={S. Zhou and J. Jia and Z. Wu and Z. Yang and Y. Wang and W. Chen and F. Meng and S. Huang and J. Shen and X. Wang}, 
year={2021}, 
month={May}, 
pages={6039-6047} }

@article{hsu2021hubert,
author = {Hsu, Wei-Ning and Bolte, Benjamin and Tsai, Yao-Hung Hubert and Lakhotia, Kushal and Salakhutdinov, Ruslan and Mohamed, Abdelrahman},
title = {HuBERT: Self-Supervised Speech Representation Learning by Masked Prediction of Hidden Units},
year = {2021},
issue_date = {2021},
publisher = {IEEE Press},
volume = {29},
issn = {2329-9290},
url = {https://doi.org/10.1109/TASLP.2021.3122291},
doi = {10.1109/TASLP.2021.3122291},
journal = {IEEE/ACM Trans. Audio, Speech and Lang. Proc.},
month = oct,
pages = {3451–3460},
numpages = {10}
}

@inproceedings{wolf2020transformers,
    title = "Transformers: State-of-the-Art Natural Language Processing",
    author = "Wolf, Thomas  and
      Debut, Lysandre  and
      Sanh, Victor  and
      Chaumond, Julien  and
      Delangue, Clement  and
      Moi, Anthony  and
      Cistac, Pierric  and
      Rault, Tim  and
      Louf, Remi  and
      Funtowicz, Morgan  and
      Davison, Joe  and
      Shleifer, Sam  and
      von Platen, Patrick  and
      Ma, Clara  and
      Jernite, Yacine  and
      Plu, Julien  and
      Xu, Canwen  and
      Le Scao, Teven  and
      Gugger, Sylvain  and
      Drame, Mariama  and
      Lhoest, Quentin  and
      Rush, Alexander",
    editor = "Liu, Qun  and
      Schlangen, David",
    booktitle = "Proceedings of the 2020 Conference on Empirical Methods in Natural Language Processing: System Demonstrations",
    month = oct,
    year = "2020",
    address = "Online",
    publisher = "Association for Computational Linguistics",
    url = "https://aclanthology.org/2020.emnlp-demos.6/",
    doi = "10.18653/v1/2020.emnlp-demos.6",
    pages = "38--45",
}

@article{lstm,
author = {Hochreiter, Sepp and Schmidhuber, J\"{u}rgen},
title = {Long Short-Term Memory},
year = {1997},
issue_date = {November 15, 1997},
publisher = {MIT Press},
address = {Cambridge, MA, USA},
volume = {9},
number = {8},
issn = {0899-7667},
url = {https://doi.org/10.1162/neco.1997.9.8.1735},
doi = {10.1162/neco.1997.9.8.1735},
journal = {Neural Comput.},
month = nov,
pages = {1735–1780},
numpages = {46}
}

@misc{pid,
      title={SPICE: Synergy and Partial Information Based Curriculum Evolution}, 
      author={Ankush Pratap Singh and Houwei Cao and Yong Liu},
      year={2026},
      eprint={2606.16639},
      archivePrefix={arXiv},
      primaryClass={cs.LG},
      url={https://arxiv.org/abs/2606.16639}, 
}
\end{document}